\documentclass[10pt,twocolumn,letterpaper]{article}

\usepackage[pagenumbers]{cvpr} 

\usepackage{times}
\usepackage{amsthm}    
\usepackage{mdframed}  
\usepackage{xcolor}  
\usepackage{color, colortbl}
\usepackage{soul}
\newmdtheoremenv[
    linecolor=white,
    backgroundcolor=blue!10,
    roundcorner=5pt,
    innertopmargin=10pt,
    innerbottommargin=10pt,
    innerrightmargin=10pt,
    innerleftmargin=10pt
]{theorem}{Theorem}

\def\name{\textsc{VisRef}\xspace}
\def\*#1{\mathbf{#1}}
\definecolor{cvprblue}{rgb}{0.21,0.49,0.74}
\definecolor{myblue}{HTML}{d8ebf8}
\definecolor{lightred}{HTML}{D33E43}
\usepackage{tabularx, booktabs}

\usepackage{amsmath,amsfonts,bm}

\def\eqref#1{equation~\ref{#1}}

\def\1{\bm{1}}

\DeclareMathAlphabet{\mathsfit}{\encodingdefault}{\sfdefault}{m}{sl}
\SetMathAlphabet{\mathsfit}{bold}{\encodingdefault}{\sfdefault}{bx}{n}

\DeclareMathOperator*{\argmax}{arg\,max}

\usepackage{url}
\usepackage{amsmath, amsthm, amssymb}
\usepackage{wrapfig}
\definecolor{mygreen}{HTML}{2DD881}
\usepackage[toc,page,header]{appendix}
\usepackage{minitoc}

\usepackage[numbers]{natbib}
\usepackage{url}
\usepackage{lipsum}
\usepackage[utf8]{inputenc} 
\usepackage[T1]{fontenc}    
\usepackage{tgtermes}
\usepackage{wrapfig}
\usepackage{tcolorbox}
\usepackage{booktabs}       
\usepackage{amsfonts}       
\usepackage{nicefrac}       
\usepackage{subcaption}
\usepackage{microtype}      
\usepackage{tabularray}
\usepackage{enumerate}
\usepackage{amsmath}
\usepackage{amsthm}

\usepackage{amssymb}
\usepackage{mathtools}
\usepackage{amsthm}
\usepackage{algorithm} 
\usepackage[noend]{algpseudocode}
\usepackage{xspace}
\usepackage{multirow}
\usepackage{enumitem}
\setlist{leftmargin=1ex}
\usepackage{lipsum}

\usepackage[normalem]{ulem}

\renewcommand{\eqref}[1]{(\ref{#1})}

\definecolor{babyblueeyes}{rgb}{0.63, 0.79, 0.95}
\usepackage{caption}
\definecolor{cvprblue}{rgb}{0.21,0.49,0.74}
\usepackage[pagebackref,breaklinks,colorlinks,allcolors=cvprblue]{hyperref}
\usepackage{pifont}

\def\paperID{8144} 
\def\confName{CVPR}
\def\confYear{2026}

\title{\name: Visual Refocusing while Thinking Improves Test-Time Scaling in Multi-Modal Large Reasoning Models}

\author{
Soumya Suvra Ghosal$^{1}$\thanks{Equal contribution. This work was done during Soumya Suvra Ghosal’s internship at AWS AI Labs.}\quad
Youngeun Kim$^{2}$\footnotemark[1]\quad
Zhuowei Li$^{2}$\quad
Ritwick Chaudhry$^{2}$\quad
Linghan Xu$^{2}$\quad
\\
Hongjing Zhang$^{2}$\quad
Jakub Zablocki$^{2}$\quad
Yifan Xing$^{2}$\quad
Qin Zhang$^{3}$\thanks{Work done while at AWS AI Labs.}
\\[2pt]
$^{1}$University of Maryland, College Park \quad
$^{2}$Amazon \quad
$^{3}$Physion Labs
\\[-2pt]
{\ttfamily\fontsize{7.5}{7}\selectfont 
sghosal@umd.edu \;
\{youngeuk,zhuoweli,ritwic,linghanx,zhhongji,jzablock,yifax\}@amazon.com \quad
qin@physionlabs.ai
}
}

\begin{document}
\maketitle

\begin{abstract}
Advances in large reasoning models have shown strong performance on complex reasoning tasks by scaling test-time compute through extended reasoning. However, recent studies observe that in vision-dependent tasks, extended textual reasoning at inference time can degrade performance as models progressively lose attention to visual tokens and increasingly rely on textual priors alone. To address this, prior works use reinforcement learning (RL)-based fine-tuning to route visual tokens or employ refocusing mechanisms during reasoning. While effective, these methods are computationally expensive, requiring large-scale data generation and policy optimization. To leverage the benefits of test-time compute without additional RL fine-tuning, we propose \name, a visually grounded test-time scaling framework. Our key idea is to actively guide the reasoning process by re-injecting a coreset of visual tokens that are semantically relevant to the reasoning context while remaining diverse and globally representative of the image, enabling more grounded multi-modal reasoning. Experiments on three visual reasoning benchmarks with state-of-the-art multi-modal large reasoning models demonstrate that, under fixed test-time compute budgets, \name\ consistently outperforms existing test-time scaling approaches by up to $6.4\%$.
\end{abstract}

\section{Introduction}
\label{sec:intro}

Multi-modal Large Reasoning Models (MLRMs)~\citep{yang2025r1, chu2025qwen, Qwen2.5-VL} have demonstrated remarkable capabilities by extending Chain-of-Thought reasoning~\citep{wei2022chain} to vision-language reasoning tasks~\citep{yue2024mmmu, lumathvista}. By generating explicit thinking traces before producing final answers, these models achieve strong performance on benchmarks requiring mathematical reasoning~\citep{lumathvista}, scientific problem-solving~\citep{yue2024mmmu}, and general multi-modal understanding~\citep{chen2024we}. However, a critical limitation emerges: for several vision-critical tasks, as these models generate longer reasoning traces, their attention to visual information progressively diminishes~\citep{chu2025qwen, yang2025look}. Visual tokens become increasingly diluted in the expanding context window, causing the model to rely more on textual priors rather than grounding its reasoning in actual image content~\citep{li2023evaluating, zhou2023analyzing}.

\begin{figure}[t]
    \centering
    \includegraphics[width=\linewidth]{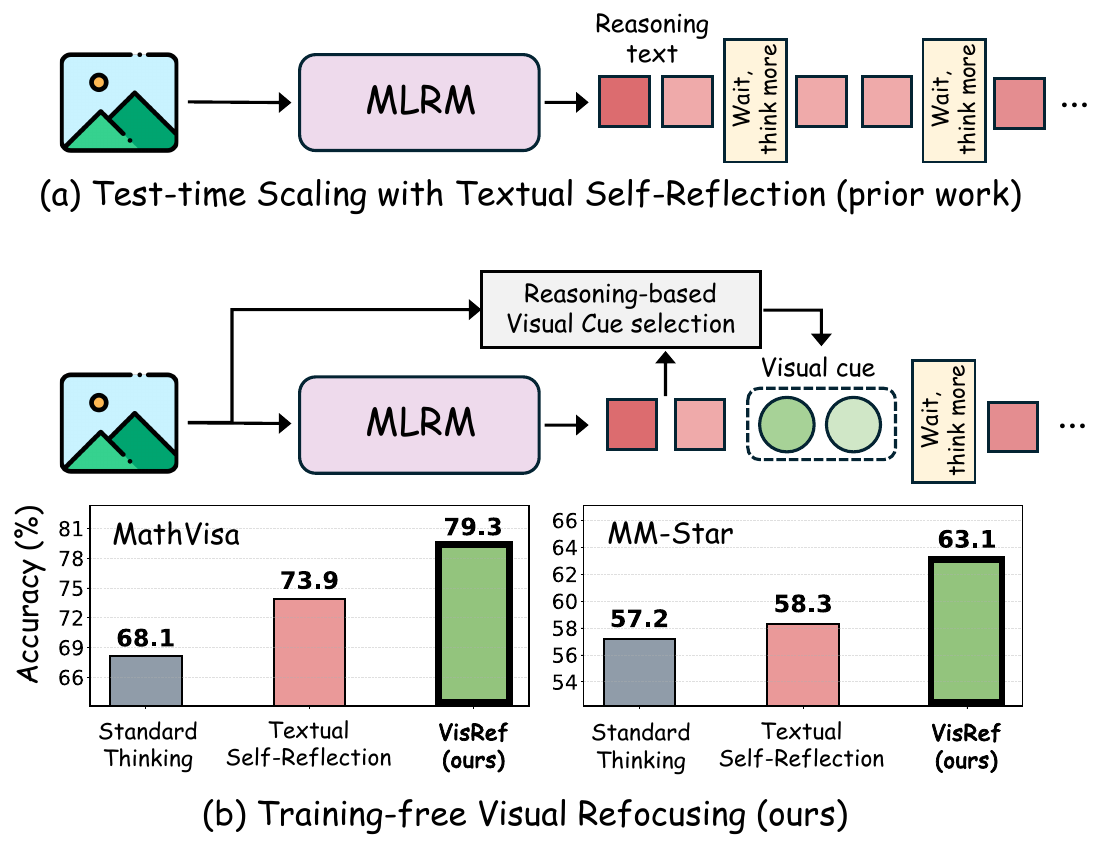}
    \caption{
       Illustration comparing our approach with prior test-time scaling methods.
    (a) Textual self-reflection-based test-time scaling \cite{muennighoff2025s1,aggarwal2025l1} extends reasoning by encouraging the model to think longer, but progressively loses grounding in the visual input as the reasoning chain grows.
    (b) Training-free Visual Refocusing (ours) dynamically selects and re-injects reasoning-relevant visual cues during inference, effectively restoring visual grounding without retraining and yielding substantial accuracy gains on the MathVista and MM-Star benchmarks with InternVL3.5-8B.
    }
    \label{fig:intro_teaser}
\end{figure}

This dilution of attention to visual tokens in MLRMs stands in stark contrast to human problem-solving. When humans reason about a multi-modal task, whether solving a geometry problem, interpreting a chart, or analyzing a diagram, they naturally alternate between examining the image and working through their reasoning, returning to verify visual details whenever uncertainty arises~\citep{barsalou2008grounded, clark2013whatever}. This interplay between perception and reasoning, where visual information grounds abstract thinking and reasoning guides visual attention, is fundamental to robust visual reasoning. Current MLRMs, however, lack this feedback loop: once visual tokens are processed initially, they progressively fade from the model's attention as textual reasoning dominates.

Recent studies have tried to address {visual dilution} through two main directions. The first employs reinforcement learning (RL) fine-tuning to teach models explicit ``look-back'' behaviors~\citep{chu2025qwen, yang2025look}. While effective, these methods are computationally expensive, and require curating large-scale annotated datasets, limiting their scalability. The second direction, test-time scaling, allocates additional inference compute to refine reasoning~\citep{snell2024scaling, muennighoff2025s1, aggarwal2025l1}. These methods generate longer reasoning chains or employ self-verification during thinking. However, existing test-time scaling approaches remain predominantly text-centric: they extend textual reasoning but fail to actively maintain visual grounding throughout the process. As the reasoning chain grows longer, visual information continues to fade, limiting their effectiveness on vision-dependent reasoning tasks~\citep{chu2025qwen}.

This gap between costly training-based solutions and ineffective text-centric test-time scaling motivates our central question: \textit{Can we restore visual grounding entirely at test time, without any retraining?} To this end, we introduce \name, a training-free framework that adaptively reinjects carefully selected visual tokens at each reasoning step, allowing the model to refocus on relevant visual content as its reasoning evolves (Fig. \ref{fig:intro_teaser}). This approach mimics the human strategy of alternating between visual examination and abstract reasoning, but does so purely at test time, requiring no specialized training data and RL fine-tuning. The core challenge is determining {which} visual tokens to reinject at each step, as naively reinjecting all tokens is computationally prohibitive and can introduce redundant information. We formulate this as an optimization problem: selecting a coreset that is both relevant to the current reasoning state and diverse in its visual coverage. We employ Determinantal Point Processes (DPPs)~\citep{kulesza2012determinantal}, which provide a principled and tractable mechanism to balance these two objectives. Furthermore, we introduce an entropy-based stopping criterion to prevent unbounded computation, terminating reasoning once the model achieves sufficient confidence.

We validate \name\ on three challenging visual-reasoning benchmarks (Section~\ref{sec:experiments}): MathVista~\citep{lumathvista}, MM-Star~\citep{chen2024we}, and MathVision~\citep{wang2024measuring}. Experiments across state-of-the-art MLRMs, including InternVL-3.5~\citep{wang2025internvl35advancingopensourcemultimodal}, Qwen-3-VL~\citep{Qwen2.5-VL}, and SAIL-VL2~\citep{yin2025sail} demonstrate consistent and significant improvements. For instance, on MathVision with SAIL-VL2, \name achieves $7.5\%$ absolute accuracy improvement over standard thinking and $5.4\%$ over textual self-reflection~\citep{muennighoff2025s1}. Furthermore, we demonstrate that \name\ scales favorably with increased test-time compute: when generating multiple parallel reasoning chains under a fixed token budget~\citep{ghosal2025does, wang2022self}, \name\ consistently achieves  superior performance for any given computational budget across all benchmarks. These results establish training-free visual refocusing as a practical and generalizable approach to maintaining visual grounding. We summarize our contributions as follows:
\begin{itemize}
    \item We propose \name, a training-free framework for adaptive visual refocusing that dynamically reinjects visual information during test-time reasoning without modifying model parameters. 
    \item We leverage a DPP-based formulation for selecting visual tokens, ensuring the selected subset is both relevant to the current reasoning state and provides diverse visual coverage in the feature space.
    \item We provide comprehensive empirical validation on challenging benchmarks (MathVista, MM-Star, MathVision) and state-of-the-art models (InternVL-3.5, Qwen3-VL, SAIL-VL2), demonstrating that \name significantly outperforms existing text-centric test-time scaling approaches.
\end{itemize}

\section{Related Works}
\label{sec:related_work}

\noindent\textbf{Multi-modal Large Reasoning Models (MLRMs).} The success of Chain-of-Thought (CoT) reasoning in LLMs~\citep{wei2022chain} spurred its adaptation to the multi-modal domain through Multimodal Chain-of-Thought (MCoT)~\citep{zhang2023multimodal, shao2024visual, fei2024video}. Initial MCoT methods relied on prompt engineering to elicit step-by-step reasoning traces. However, these short, reactive chains often proved insufficient for complex, real-world tasks requiring long-horizon planning~\citep{zhang2024mme, zhao2024marco, yue2024mmmu}. To address this gap, recent research has shifted toward using reinforcement learning~\citep{guo2025deepseek} to instill more deliberate and methodologically structured reasoning processes. This paradigm shift, notably influenced by work like DeepSeek-R1~\citep{guo2025deepseek}, has inspired a new generation of MLRMs designed for deeper reasoning~\citep{yang2025r1, deepmind2024gemini25, huang2025vision, peng2025lmm, thawakar2025llamav, chen2025sftrlearlyinvestigation, deng2025openvlthinker, yao2024mulberry, xu2024llava, team2025kimi, wang2025internvl35advancingopensourcemultimodal, Qwen2.5-VL}.

\vspace{0.2cm}
\noindent\textbf{Test-time Scaling in Reasoning Models.} Recent work by \citet{muennighoff2025s1} introduced the concept of budget forcing to replicate the test-time scaling behavior observed in o1 models~\citep{openo1}. Another recent approach, L1~\citep{aggarwal2025l1}, proposed length-controlled policy optimization, providing precise control over the length of the reasoning trace during generation. \citet{yang2025thinkingoptimalscalingtesttimecompute} introduced a thinking-optimal scaling strategy, training models to adapt dynamically to different levels of reasoning effort depending on the test-time compute budget. Recently, a lot of studies have also focused on fine-tuning models to think efficiently according to task complexity~\citep{arora2025traininglanguagemodelsreason, fang2025thinkless, zhang2025continue, jiang2025think, liang2025thinkswitcher, zhang2025adaptthink, huang2025adactrl}. However, a growing body of evidence indicates that simply extending the thinking process at test time can lead to oscillatory performance~\citep{ghosal2025does, wu2025more, hassid2025don}. 
In the multimodal setting, \citet{chu2025qwen} observed a similar trend: as reasoning length increases, attention to visual tokens degrades, harming visual grounding.

\vspace{0.2cm}
\noindent\textbf{Visual refocusing in MLRMs.}
Recent works~\citep{chu2025qwen, yang2025look, zhang2025mllms} has emphasized the need to refocus on visual tokens during multi-modal reasoning, as extended inference often leads to degradation of visual grounding. \citet{chu2025qwen} address this by fine-tuning models with explicit ``look-back'' or ``reflect'' mechanisms that copy or route visual tokens during inference. Similarly, \citet{yang2025look} proposed an implicit refocusing scheme using specialized supervision to encourage models to revisit image context mid-reasoning. Another emerging line of research explores enhancing multi-modal models through the integration of external visual tools. Several approaches incorporate zoom, cropping, or agent-based utilities~\cite{su2025pixel, zhang2025chain, huang2025visualtoolagent, zhang2025mllms}, or leverage advanced vision modules through supervised fine-tuning or reinforcement learning to enable tool-using behaviors~\cite{chung2025don, zhang2025chain,geng2025webwatcher}. While above-mentioned methods successfully enhance visual grounding, they typically require specialized dataset construction, model retraining, tool-calling, or architectural modifications. In contrast, our framework, \name, enables efficient and adaptive visual refocusing purely at test time with minimal overhead, making it plug-and-play for any pre-trained MLRM.

\section{Preliminaries}
\label{sec:prelims}

\noindent\textbf{Mathematical formulation of thinking process.} Formally, an MLRM-generated thinking process can be expressed as:
\begin{equation}
\label{eq:standard_thinking}
    x_{\text{input}} \;\rightarrow\; z\;\rightarrow\; y ,
\end{equation}
where $x_{\text{input}} = [I, T]$  denotes the input, with ${I} \in \mathcal{I}$ the visual input (\eg, image) and ${T} = (t_{1}, t_{2}, \ldots, t_{P})$ the textual prompt consisting of $P$ tokens ($t_{i} \in \mathcal{M}$ for vocabulary $\mathcal{M}$). The model first produces a reasoning (or thinking) trace $z \sim {\pi}_\theta(\cdot|x_{\text{input}})$ and then a final answer $y \sim {\pi_\theta}(\cdot|x_{\text{input}}, z)$. The RL objective for training reasoning models is:
\begin{align}
    \max_{\theta} \mathop{\mathbb{E}}_{\substack{x_{\text{input}}, z \sim \pi_\theta(\cdot|x_{\text{input}}), \\ y \sim \pi_\theta(\cdot|x_{\text{input}}, z)}} [R(x_{\text{input}},y)]   ,
\end{align}
where $\pi_{\theta}$ is the parameterized model and $R(x_{\text{input}}, y)$ represents the true reward function (\eg, an indicator to check if $y$ is correct or not), which is obtained once the policy generates the final response $y$.

\vspace{0.2cm}
\noindent\textbf{Textual self-reflection in reasoning models.} Textual self-reflection~\citep{muennighoff2025s1} extends the thinking process as:
\begin{equation*}
    x_{\text{input}} \rightarrow z_1 \rightarrow z_2 \rightarrow \cdots \rightarrow z_k \rightarrow y,
\end{equation*}
where, given the prompt $x_{\text{input}}$, the model first generates an initial reasoning step $z_{1} \sim \pi_{\theta}(\cdot \mid x_{\text{input}})$. Rather than producing the final answer immediately, the model is prompted to continue reasoning using special instruction tokens (e.g., “Wait”, “Think more”), denoted by $c$. Subsequent reasoning steps are sampled iteratively as $z_{t} \sim \pi_{\theta}(\cdot \mid x_{\text{input}}, z_{1:t-1}, c)$ for $t = 2, \ldots, k$, until the model generates final response as $y \sim \pi_{\theta}(\cdot \mid x_{\text{input}}, z_{1:k}, c)$. For brevity, we omit explicit mention of $c$ in the conditioning and denote the thinking traces in the condition as $y \sim \pi_\theta(\cdot|x_{\text{input}}, z_{1:k})$.


\section{Proposed Framework}

\noindent\textbf{Problem Setup.} Recent studies~\citep{chu2025qwen, yang2025look} have observed that in vision-dependent reasoning tasks~\citep{lumathvista, yue2024mmmu}, scaling test-time compute by extending textual reasoning often leads to a  dilution of visual information, weakening the influence of visual tokens and inducing visual hallucinations~\citep{li2025hidden, zhou2023analyzing, favero2024multi}. To mitigate this issue, prior works~\citep{yang2025look, chu2025qwen} employed RL fine-tuning~\citep{guo2025deepseek} to enable models to autonomously decide {when} and {how} to refocus on visual input during reasoning. While empirically effective, RL-based approaches are computationally intensive and require costly data curation. This motivates a central question: \emph{Can we achieve adaptive visual refocusing entirely at test time, without additional fine-tuning?} 

We address this challenge by introducing \name, a training-free framework that dynamically reintegrates visual information during thinking. Our key insight is to augment the textual reasoning trace by adaptively reinjecting relevant visual tokens at each step, ensuring the model maintains grounded visual context throughout its reasoning process. 

Formally, given an image–text input $x_{\text{input}} = [I, T]$, let $\mathcal{V}=\{v_1, v_2, \cdots, v_N\}$ denote the set of $N$ visual token embeddings extracted from image $I$, where each $v_i \in \mathbb{R}^d$ is a $d$-dimensional vector. At reasoning step $k$, the {visual-integrated reasoning trajectory} can be expressed as:
\begin{equation}
\label{eq:visual_traj}
    \tau_{1:k} = \big\{(z_1, V_1), (z_2, V_2), \ldots, (z_k, V_k)\big\},
\end{equation}
where $z_i$ denotes the textual reasoning step at step $i$ and $V_i \subseteq \mathcal{V}$ represents the subset of visual tokens reinjected at that step. The model continues the reasoning process until a stopping condition is met and generates the final answer as $y \sim \pi_{\theta}(\cdot \mid x_{\text{input}}, \tau_{1:k})$. The stopping criterion is important as indefinitely reinjecting tokens may degrade performance due to overthinking~\citep{snell2024scaling, ghosal2025does}, while premature stopping can yield incomplete reasoning and incorrect predictions. 

\begin{figure*}[t]
    \centering
    \includegraphics[width=0.87\linewidth]{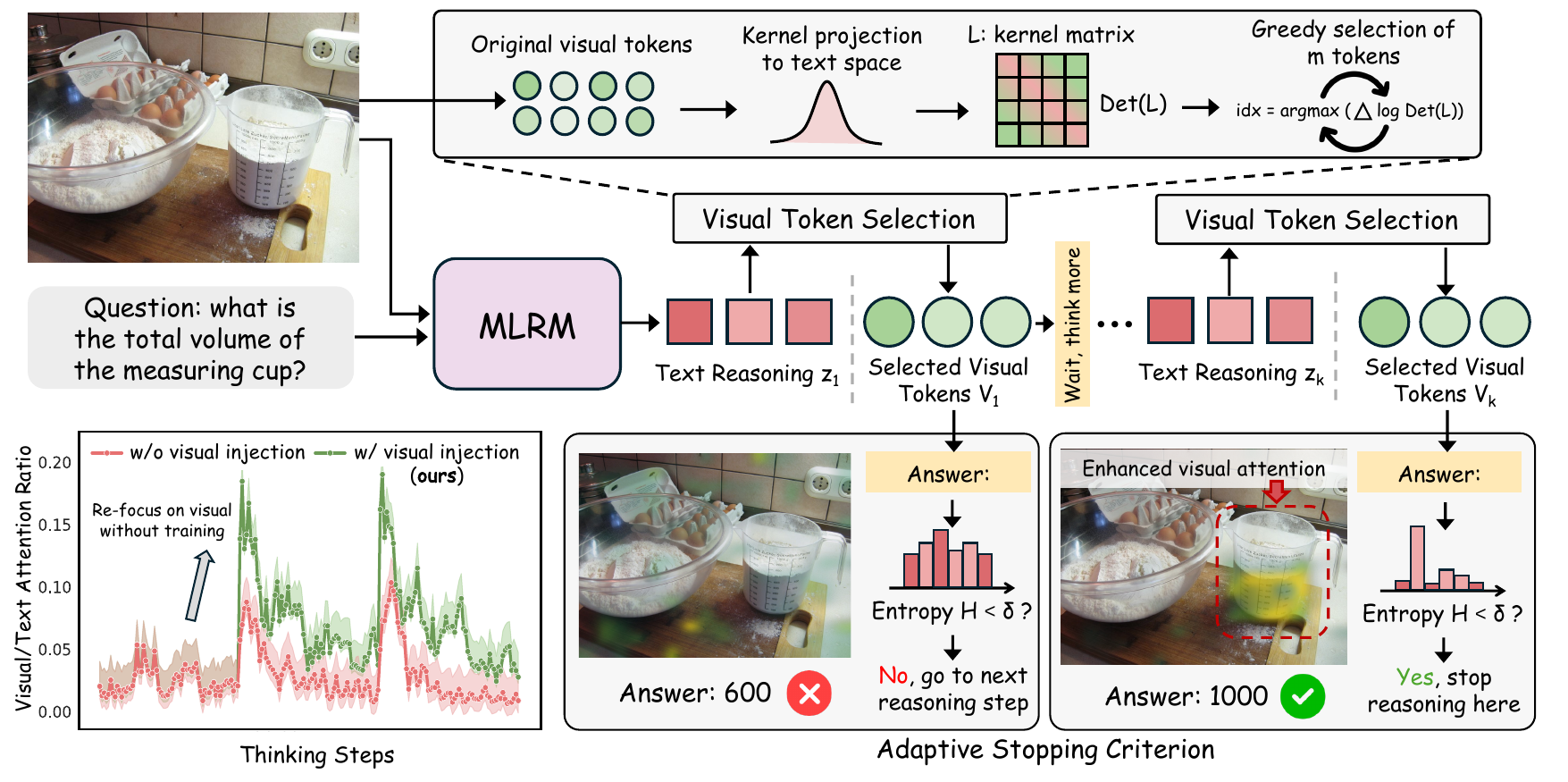}
    \vspace{-3mm}
    \caption{
        \textbf{Overview of \name.} Given an image-text input, \name enables multi-modal large reasoning models (MLRMs) to maintain visual grounding throughout the reasoning process without retraining. At each reasoning step, the model projects visual tokens into the textual reasoning subspace and selects a subset of reasoning-relevant tokens via a determinantal point process (DPP)-based criterion. The selected tokens are then reinjected to guide subsequent reasoning. This iterative process continues until the entropy of the model's answer distribution falls below a confidence threshold $\delta_{\mathrm{entropy}}$, forming an adaptive stopping criterion. The bottom-left plot shows the visual-to-text attention ratio across reasoning steps. As shown, our method maintains a higher level of visual attention (\textcolor{ForestGreen}{green} curve) by reinjecting a carefully selected coreset of visual tokens, whereas text-based self-reflection (\textcolor{red}{red} curve) exhibits a more rapid decline in visual attention, consistent with prior observations~\citep{yang2025look, chu2025qwen}. Example shown uses InternVL-3.5-8B on MathVista.}
    \vspace{-3mm}
    \label{fig:main_figure}
\end{figure*}

\subsection{Our Approach: \name}
\label{subsub:our_approach}

Enabling autonomous visual refocusing during reasoning presents two key challenges: (1) \textit{visual token selection}: identifying which subset of visual tokens to re-inject at each step, and (2) \textit{stopping criterion}: determining when to terminate reasoning and generate the final answer. We address challenge 1 in \ref{subsubsec:visual_selection} and challenge 2 in \ref{subsubsec:stopping}. The overall method is illustrated in Fig. \ref{fig:main_figure}.

\subsubsection{Identifying Optimal Visual Token Subsets}  
\label{subsubsec:visual_selection}

A naive solution to mitigate visual token dilution would be reinjecting all $N$ visual tokens at every reasoning step. However, this strategy is computationally prohibitive in practice, significantly increasing context length and inference latency. For example, using InternVL-3.5-8B~\citep{wang2025internvl35advancingopensourcemultimodal} on MathVista~\citep{lumathvista}, the number of visual tokens per image averages approximately $3\times$ the text tokens generated in each reasoning step (e.g., $\sim$1,772 visual tokens vs. $\sim$615 text tokens), and appending all visual tokens leads to a $2.3\times$ increase in inference latency compared to text-only reasoning. Hence, to address this computational bottleneck, we seek to select a {coreset} of visual tokens $V_k \subseteq \mathcal{V}$ at each reasoning step $k$ that maximizes the expected reward of the final answer. Formally, the optimal coreset satisfies:

\begin{footnotesize}
\begin{align}
\label{eq:ours_1}
    V_k^* = \arg\max_{V_k \subseteq \mathcal{V}} \;
    \mathop{\mathbb{E}}_{\substack{y \sim \pi_\theta(\cdot \mid x_{\text{input}}, \tau_{1:K}), \\
        z_{k+1:K} \sim \pi_\theta(\cdot \mid x_{\text{input}}, \tau_{1:k-1}, z_k, V_k)
    }}
    \big[ R(x_{\text{input}}, y) \big],
\end{align}
\end{footnotesize}

\noindent where $\tau_{1:k-1} = \{(z_1, V_1), \ldots, (z_{k-1}, V_{k-1})\}$ denotes the visually integrated reasoning trajectory up to step $k-1$. The objective in Equation~\ref{eq:ours_1} aims to find the visual token subset $V_k$ that maximizes the expected reward of the final answer $y$, marginalized over all possible future reasoning paths. However, directly optimizing Equation~\ref{eq:ours_1} is intractable at test time, as it requires (i) access to the true reward function $R$, which is only available after generating the final answer, and (ii) enumerating over the exponentially large space of visual token subsets and future trajectories.

\vspace{0.2cm}
\noindent\textbf{Tractable test-time optimization.} To circumvent the challenge of directly optimizing Equation~\ref{eq:ours_1}, we reformulate the problem by introducing a simplifying assumption that leads to a tractable test-time objective. Specifically, we adopt a {Markov assumption}, positing that the utility of $V_k$ at step $k$ depends primarily on the current reasoning state $z_k$ rather than the entire reasoning history $\tau_{1:k-1}$. This is reasonable since $z_k$ is generated autoregressively conditioned on $\tau_{1:k-1}$, thus implicitly encoding the relevant historical context. With this assumption, we replace the intractable expected reward in Equation~\ref{eq:ours_1} with a tractable proxy test-time objective:
\begin{align}
\label{eq:ours_2}
    \widetilde{V}_k = \arg\max_{V_k \subseteq \mathcal{V}} \;
    \mathcal{J}(V_k \mid x_{\text{input}}, z_k),
\end{align}
where $\mathcal{J}(V_k \mid x_{\text{input}}, z_k)$ is a scoring function that helps select the best visual tokens in $V_k$.

\vspace{0.2cm}
\noindent\textbf{Designing the scoring function $\mathcal{J}$.} Intuitively, we seek a coreset $V_k$ whose tokens are not only relevant to the current reasoning step $z_k$ but also maximally diverse to ensure maximum coverage of the visual content and avoid redundancy. Geometrically, this corresponds to finding tokens whose feature representations, when projected into the subspace defined by $z_k$, span the largest volume -- ensuring relevance through projection and diversity through volume maximization~\citep{taskar2013determinantal, civril2009selecting, kulesza2012determinantal}.

To formalize this intuition, we employ Determinantal Point Processes (DPPs)~\citep{taskar2013determinantal, kulesza2012determinantal}, a probabilistic framework that naturally captures both relevance and diversity through a kernel-based determinant. Let $z_k = \{z^{(1)}_k, z^{(2)}_k, \ldots, z^{(T_k)}_k\} \in \mathbb{R}^{T_k \times d}$ denote the $T_k$ text token embeddings in reasoning state $z_k$. We capture the reasoning subspace geometry via:
\begin{align}
    M_k = \sum_{i=1}^{T_k} z^{(i)}_k(z^{(i)}_k)^{\top}.
\end{align}
We then define a positive semi-definite similarity kernel that measures how similar any pair of visual tokens $v_i, v_j$ appear when projected into this textual subspace $L_k(v_i, v_j) = \phi_k(v_i)^{\top}\phi_k(v_j)$ where $\phi_k(v) = M_k^{1/2}v$ embeds the visual token $v$ into a feature space aligned with the geometry of the reasoning subspace spanned by $z_k$. Using this kernel, we define our scoring function as the determinant of the kernel matrix restricted to subset $V_k$:
\begin{align}
\label{eq:scoring_det}
    \mathcal{J}(V_k \mid x_{\text{input}}, z_k) = \det(L_k^{V_k}),
\end{align}
where $L_k^{V_k} \in \mathbb{R}^{|V_k| \times |V_k|}$ is the kernel matrix restricted to the subset $V_k$, with entries $[L_k^{V_k}]_{ij} = L_k(v_i, v_j)$ for $v_i, v_j \in V_k$. The determinant naturally balances relevance and diversity: it increases when tokens are individually relevant to $z_k$ (high diagonal values) while penalizing redundancy through the correlation structure (off-diagonal terms). By jointly considering Equations~\ref{eq:ours_2} and~\ref{eq:scoring_det}, our final optimization becomes:
\begin{align}
\label{eq:final_opt}
    \widetilde{V}_k = \arg\max_{V_k \subseteq \mathcal{V}} \; \det(L_k^{V_k}).
\end{align}

\begin{algorithm}[t]
\caption{\name: Visual Refocusing while Thinking}
\label{alg:main}
\begin{algorithmic}[1]
\Require Image-text input $x_{\text{input}} = [I, T]$, visual tokens $\mathcal{V} = \{v_1, \ldots, v_N\}$, token budget $m$, entropy threshold $\delta_{\text{entropy}}$, maximum steps $K_{\text{max}}$
\Ensure Final answer $y$
\State Initialize $k \gets 1$, $\tau \gets \emptyset$
\While{$k \leq K_{\text{max}}$}
    \State \textcolor{gray}{{\footnotesize// Generate reasoning step}}
    \State Sample $z_k \sim \pi_\theta(\cdot \mid x_{\text{input}}, \tau_{1:k-1})$
    \State {\footnotesize\textcolor{gray}{// Select relevant and diverse visual tokens (Section~\ref{subsubsec:visual_selection})}}
    \State Compute text subspace: $M_k = \sum_{j=1}^{T_k} z_k^{(j)} (z_k^{(j)})^\top$
    \State Define kernel: $L_k(v_i, v_j) = v_i^\top M_k v_j$
    \State $\widetilde{V}_k \gets$ Greedy selection of $m$ tokens via Eq.~\eqref{eq:greedy_selection}
    \State {\footnotesize\textcolor{gray}{// Update trajectory}}
    \State $\tau_{1:k} \gets \tau_{1:k-1} \cup \{(z_k, \widetilde{V}_k)\}$
    \State {\footnotesize\textcolor{gray}{// Check stopping criterion (Section~\ref{subsubsec:stopping})}}
    \State $H_k \gets -\mathop{\mathbb{E}}_{y \sim \pi_\theta} [\log \pi_\theta(y \mid x_{\text{input}}, \tau_{1:k})]$
    \If{$H_k < \delta_{\text{entropy}}$}
        \State \textbf{break}
    \EndIf
    \State $k \gets k + 1$
\EndWhile
\State {\footnotesize\textcolor{gray}{// Generate final answer}}
\State Sample $y \sim \pi_\theta(\cdot \mid x_{\text{input}}, \tau_{1:k})$
\State \Return $y$
\end{algorithmic}
\end{algorithm}

\vspace{0.2cm}
\noindent\textbf{Why maximizing the determinant works?} Maximizing $\log \det(L_k^{V_k})$ in Equation~\eqref{eq:final_opt} explicitly balances relevance and diversity. For each visual token $v_i$, we define its relevance to the current reasoning state $z_k$ as
\begin{align}
    r_i^2 = \phi_k(v_i)^\top \phi_k(v_i) 
          = v_i^\top M_k v_i 
          = \sum_{j = 1}^{T_k} (v_i^{\top} z_k^{(j)})^2,
    \label{eq:relevance_term}
\end{align}
which measures the alignment between token $v_i$ and the text-conditioned representation $z_k$. We also define a normalized diversity kernel $[\bar{L}_k]_{ij} = [L_k]_{ij}/(r_i r_j)$, so that $\forall V_k$, the kernel matrix factorizes as $L_k^{V_k} = \mathrm{Diag}(r_{V_k}) \, \bar{L}_k^{V_k} \, \mathrm{Diag}(r_{V_k})$, yielding (full derivation in Appendix):
\begin{align}
    \log \det(L_k^{V_k})
    = \underbrace{\sum_{v_i \in V_k} \log(r_i^2)}_{\text{relevance}}
    + \underbrace{\log \det(\bar{L}_k^{V_k})}_{\text{diversity}}.
    \label{eq:ours_decomp}
\end{align}
This decomposition reveals that our objective naturally balances: (1) the \emph{relevance term}, measuring the overall alignment of selected visual tokens with the textual context, and (2) the \emph{diversity term}, ensuring selected tokens are mutually dissimilar ensuring maximum visual coverage.

\vspace{0.2cm}
\noindent\textbf{Efficient inference-time solution.}
The combinatorial optimization in Equation~\eqref{eq:final_opt} is NP-hard~\citep{ko1995exact}. 
Hence, following prior works~\citep{chen2018fastgreedymapinference, ye2023compositional}, we approximate the solution via a greedy selection algorithm. 
We introduce a token budget $m$ that controls the number of visual tokens selected at each reasoning step, enforcing $|V_k| = m$.
Starting from $V_{k}^{(0)} = \emptyset$, at iteration $i$ we select the token with maximum marginal gain:
\begin{equation}
v_{k,i}=\argmax_{v\in\mathcal{V}\setminus V_{k}^{(i-1)}}\!
\left[\log\bigg(\frac{\det(L_k^{V_{k}^{(i-1)}\cup\{v\}})}{\det(L_k^{V_{k}^{(i-1)}})}\bigg)\right]
\label{eq:greedy_selection}
\end{equation}
and update $V_{k}^{(i)} \leftarrow V_{k}^{(i-1)} \cup \{v_{k,i}\}$.
We terminate after $m$ iterations, setting $\widetilde{V}_k \leftarrow V_k^{(m)}$.

\subsubsection{Adaptive Stopping Criterion}
\label{subsubsec:stopping}

Beyond selecting the optimal visual tokens, the framework must also determine when to terminate reasoning and produce the final answer. For this, we propose an adaptive stopping criterion based on the model's predictive confidence. At each reasoning step $k$, given the current visual-integrated trajectory $\tau_{1:k} = \{(z_1, V_1), \ldots, (z_k, V_k)\}$, we assess the model's certainty by measuring the entropy of its response distribution, defined as $H_k = -\mathbb{E}_{y \sim \pi_{\theta}(\cdot \mid x_{\text{input}}, \tau_{1:k})} [\log \pi_{\theta}(y \mid x_{\text{input}}, \tau_{1:k})]$. 
A low entropy $H_k < \delta_{\text{entropy}}$, where $\delta_{\text{entropy}}$ is a threshold hyperparameter, signals that the model has converged to a confident answer distribution where further reasoning is unlikely to yield improvement. Conversely, high entropy indicates uncertainty and the potential benefit of additional reasoning steps with visual refocusing. To bound computation, we enforce a maximum number of reasoning steps, $K_{\text{max}}$. This criterion naturally adapts to problem difficulty: simpler questions reach low entropy quickly, while complex problems utilize extended reasoning before achieving sufficient confidence. After termination of the reasoning trajectory, the final answer is generated as: $y \sim \pi_\theta(\cdot \mid x_{\text{input}}, \tau_k)$.

The complete framework, combining our visual token selection (Section~\ref{subsubsec:visual_selection}) with the adaptive stopping criterion (Section~\ref{subsubsec:stopping}), is summarized in Algorithm~\ref{alg:main}.

\section{Experiments}
\label{sec:experiments}


\begin{table}[!t]
\centering
\caption{
\textbf{Evaluation on visual reasoning benchmarks.}
We evaluate \name across three visual reasoning benchmarks.
To ensure a fair comparison, all methods adopt the adaptive stopping criterion described in Section~\ref{subsubsec:stopping}.  
For brevity, we denote \textit{Standard Thinking} as \textbf{ST}, and \textit{Textual Self-Reflection}~\citep{muennighoff2025s1} as \textbf{TSR}.  
All results are reported in accuracy (\%), and the numbers in parentheses indicate the performance gain over the ST baseline.
}
\label{tab:main_results_adaptive}
\resizebox{\linewidth}{!}{
\begin{tabular}{l|l|ccc}
\toprule
\multirow{1}{*}{Model} & \multirow{1}{*}{Method} 
& {\textbf{MathVision}} 
& {\textbf{MathVista}} 
& {\textbf{MM-Star}} \\
\midrule

\multirow{3}{*}{{InternVL3.5-8B}} 
& ST (Baseline) & 39.2  & 68.1  & 57.2  \\
& TSR~\citep{muennighoff2025s1} & 40.1  & 73.9  & 58.3 \\
& \cellcolor{myblue!55}\textbf{VisRef (Ours)} 
  & \cellcolor{myblue!55}\textbf{44.6 (\textcolor{ForestGreen}{+4.5})} 
  & \cellcolor{myblue!55}\textbf{79.3 (\textcolor{ForestGreen}{+5.4})} 
  & \cellcolor{myblue!55}\textbf{63.1 (\textcolor{ForestGreen}{+4.8})} \\
\midrule

\multirow{3}{*}{{Qwen3-VL-8B}} 
& ST (Baseline) & 53.8  & 74.1  & 66.5  \\
& TSR~\citep{muennighoff2025s1} & 54.3  & 74.2  & 65.9  \\
& \cellcolor{myblue!55}\textbf{VisRef (Ours)} 
  & \cellcolor{myblue!55}\textbf{56.6 (\textcolor{ForestGreen}{+2.3})}  
  & \cellcolor{myblue!55}\textbf{77.1 (\textcolor{ForestGreen}{+2.9})} 
  & \cellcolor{myblue!55}\textbf{69.1 (\textcolor{ForestGreen}{+3.2})} \\
 \midrule

\multirow{3}{*}{{SAIL-VL2-8B}} 
& ST  (Baseline)& 29.8  & 73.1 & 47.7  \\
& TSR~\citep{muennighoff2025s1} & 31.9 & 73.8  & 48.9 \\
& \cellcolor{myblue!55}\textbf{VisRef (Ours)} 
  & \cellcolor{myblue!55}\textbf{37.3 (+\textcolor{ForestGreen}{5.4})} 
  & \cellcolor{myblue!55}\textbf{78.2 (+\textcolor{ForestGreen}{4.4})} 
  & \cellcolor{myblue!55}\textbf{55.3 (+\textcolor{ForestGreen}{6.4})} \\
\bottomrule
\end{tabular}}
\end{table}

\noindent\textbf{Experimental Setup}
\label{sec:exp_setup}.
We evaluate \name on three visual reasoning benchmarks—MathVista~\cite{lu2023mathvista} (\textit{testmini}, 1{,}000 problems), MathVision~\cite{wang2024measuring} (304 competition-style problems), and MM-Star~\cite{chen2024we} (1{,}500 vision-dependent questions)—using three state-of-the-art MLRMs in their reasoning (\textit{thinking}) mode: InternVL3.5-8B~\cite{wang2025internvl3}, SAIL-VL2-Thinking~\cite{yin2025sail}, and Qwen-3-VL-8B-Thinking~\cite{Qwen2.5-VL}. We compare against (i) Standard Thinking (ST; Eq.~\ref{eq:standard_thinking}) and (ii) Textual Self-Reflection (TSR)~\citep{muennighoff2025s1}; unless otherwise stated, we set the adaptive stopping entropy threshold $\delta_{\text{entropy}}{=}0.25$, visual token budget $m=\lfloor 0.3|\mathcal{V}|\rfloor$, and maximum reasoning steps $K_{\text{max}}{=}10$. We report test accuracy by checking whether each model’s final prediction $y$ matches the ground-truth answer $y^*$ for each image-text input. A detailed description is provided in the Appendix.

\begin{figure*}
    \centering
    \includegraphics[width=0.90\textwidth]{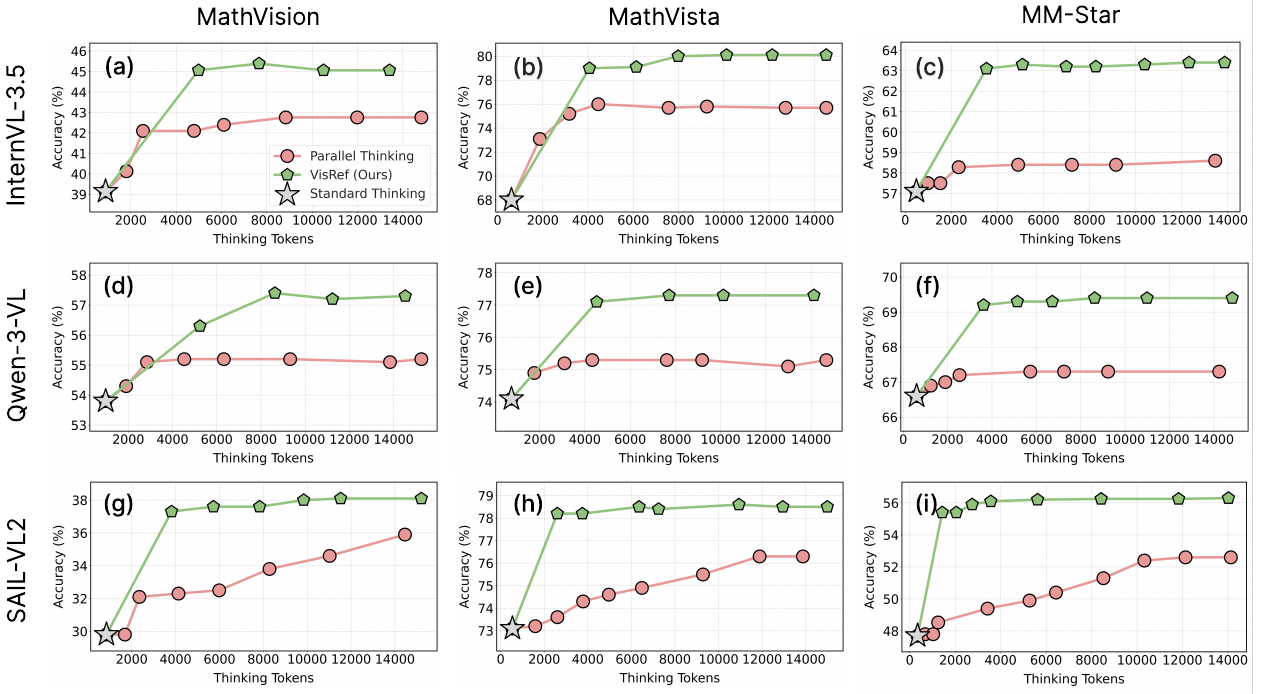}
    \vspace{-2mm}
\caption{ \textbf{Test-time scaling of \name.} We evaluate the test-time scaling behavior of \name by generating multiple parallel visual-integrated reasoning chains under a fixed token budget. Results are shown across three benchmarks (MathVision, MathVista, and MM-Star) and three MLRMs: InternVL-3.5-8B (first row), Qwen-3-VL-8B (second row), and SAIL-VL2 (third row). The star marker (\ding{73}) denotes standard thinking—the baseline with no additional test-time compute. Parallel thinking~\citep{ghosal2025does, wang2022self} generates multiple parallel chains-of-thought without visual refocusing. Across all models and benchmarks, \name consistently achieves superior accuracy for any given computational budget.  }
\label{fig:parallel_results}
\vspace{-1mm}
\end{figure*}

\noindent\textbf{Evaluation Results.} Table~\ref{tab:main_results_adaptive} presents evaluation results across three visual reasoning benchmarks and three MLRMs comparing \name with standard thinking (ST), and textual self-reflection (TSR)~\citep{muennighoff2025s1}. To ensure fair comparison, we employ the same adaptive stopping criterion (Section~\ref{subsubsec:stopping}) for textual self-reflection as well. First, we observe that textual self-reflection provides inconsistent improvements over standard thinking, with gains ranging from $0.1\%$ to $2.1\%$ on most benchmarks but notably degrading performance by $0.6\%$ on MM-Star with Qwen-3-VL-8B. This inconsistency suggests that purely textual reasoning extension offers limited and unreliable benefits for vision-dependent tasks.
In contrast, \name consistently outperforms both baselines across all settings. Using InternVL-3.5-8B, \name achieves substantial improvements of $5.4\%$, $11.2\%$, and $5.9\%$ over standard thinking on MathVision, MathVista, and MM-Star, respectively. Compared to textual self-reflection, \name gains an additional $4.5\%$ and $5.4\%$ on MathVision and MathVista respectively. Similar improvements are observed with Qwen-3-VL-8B $(2.8\%, 3.0\%, 2.6\%)$ and SAIL-VL2-8B $(7.5\%, 5.1\%, 7.6\%)$, demonstrating the generalizability of our approach across different model architectures. These results highlight that selective visual token reinjection effectively maintains visual grounding throughout reasoning.

\noindent\textbf{Understanding test-time scaling behavior.} Recent studies~\citep{ghosal2025does, feng2025characterizes} show that given a fixed test-time thinking token budget $B$, an efficient utilization approach is to generate multiple parallel chains of thought. Following this, we evaluate how \name scales with increased test-time compute by generating $\mathcal{C}$ parallel visual-integrated thinking traces for a given image-text pair $x_{\text{input}}$: $\tau^{(i)} \sim \pi_{\theta}(\cdot\mid x_{\text{input}})$, $i = 1, 2, \ldots, \mathcal{C}$, subject to $\sum_{i=1}^{\mathcal{C}} |\tau^{(i)}| \leq B$. Each trace $\tau^{(i)}$ is sampled independently using Algorithm~\ref{alg:main}. For each reasoning trace $\tau^{(i)}$, we generate a final answer $y^{(i)} \sim \pi_\theta(\cdot \mid \tau^{(i)}, x_{\text{input}})$, yielding a candidate set $\mathcal{Y}=\{y^{(1)}, y^{(2)}, \ldots, y^{(\mathcal{C})}\}$. Following self-consistency aggregation~\citep{wang2022self}, we select the final output via majority voting: $\widetilde{y} = \argmax_{y \in \mathcal{Y}} \sum_{i=1}^\mathcal{C} \mathbb{I}[y^{(i)} = y],$ where $\mathbb{I}$ represents the indicator function.

Figure~\ref{fig:parallel_results} presents the test-time scaling behavior of \name across three visual reasoning benchmarks—MathVision~\citep{wang2024measuring}, MathVista~\citep{lumathvista}, and MM-Star~\citep{chen2024we}—and three MLRMs-- InternVL-3.5-8B (first row), Qwen-3-VL-8B (second row), and SAIL-VL2 (third row). The star marker (\ding{73}) denotes the baseline with no additional test-time compute (standard thinking), while each successive circle to the right represents increasing test-time token budget. We compare \name against parallel thinking~\citep{ghosal2025does, wang2022self}, which samples multiple text-only reasoning trajectories without visual refocusing. Across all benchmarks and models, \name consistently achieves superior accuracy for any given computational budget. On MM-Star using InternVL-3.5-8B with a budget of $14K$ thinking tokens, \name achieves around 6\% higher accuracy compared to parallel thinking, demonstrating the benefit of maintaining visual grounding throughout the reasoning process.

\begin{figure}[!t]
    \centering
    \includegraphics[width=0.9\linewidth]{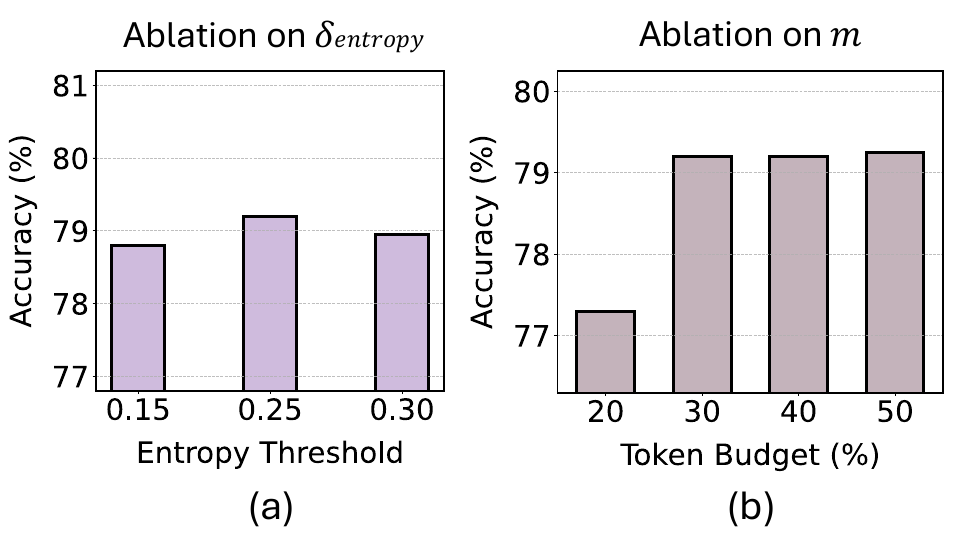}
    \vspace{-4mm}
    \caption{\textbf{Ablation study of hyper-parameters.} 
(a) We visualize the ablation results of the entropy threshold $\delta_{\mathrm{entropy}}$ used as the stopping criterion. Although the accuracy does not vary significantly across different thresholds, our evaluation shows that $\delta_{\mathrm{entropy}} = 0.25$ achieves the best balance between accuracy and inference efficiency.
(b) We show the ablation of the token budget $m$ (fraction of visual tokens selected) on {MathVista} using {InternVL-3.5-8B}. Accuracy improves from 76.1\% to 79.2\% as $m$ increases from 20\% to 30\% but plateaus for $m \geq 30\%$.
} 
    \vspace{-3mm}
    \label{fig:ablation}
\end{figure}

\begin{figure*}[t!]
    \centering
    \includegraphics[width=\linewidth]{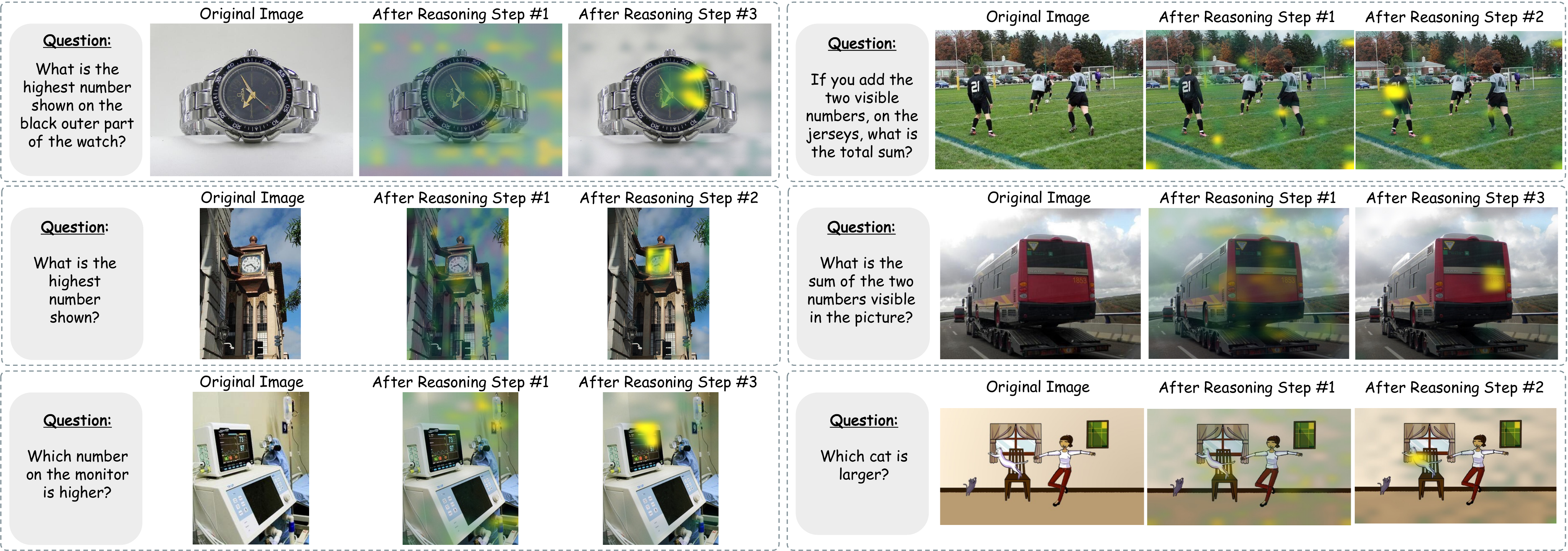}
    \vspace{-5mm}
    \caption{\textbf{Attention Visualization.} 
Attention maps show how \name progressively refocuses on relevant visual regions during multi-step reasoning. Initially, the attention maps are noisy. With visual reinjection, \name  reinforces grounding on task-critical objects, leading to more accurate visual reasoning.
} 
    \vspace{-2.5mm}
    \label{fig:visualization}
\end{figure*}

\noindent\textbf{Comparison with training-based methods.}
Table~\ref{tab:lookback_allbench} compares \name with Look-Back~\cite{yang2025look} using InternVL-3.5-8B. While Look-Back achieves strong performance through RL fine-tuning, \name attains competitive results \textit{entirely without training}. Moreover, combining Look-Back~[1] with \name yields the best performance across all benchmarks, demonstrating that our approach is orthogonal to training-based methods and can provide complementary gains. Crucially, \name is training-free and can be immediately applied to any pretrained MLRM, whereas Look-Back requires fine-tuning for 60 GPU hours with A6000.

\noindent\textbf{Comparison with training-based methods.}
Table~\ref{tab:lookback_allbench} compares \name with Look-Back~[1] using InternVL-3.5-8B. While Look-Back achieves strong performance through RL fine-tuning, \name attains competitive results \textit{entirely without training}. Moreover, combining Look-Back~[1] with \name yields the best performance across all benchmarks, demonstrating that our approach is orthogonal to training-based methods and can provide complementary gains. Crucially, \name is training-free and can be immediately applied to any pretrained MLRM, whereas Look-Back~[1] requires fine-tuning for 60 GPU hours with A6000.

\begin{table}[t]
    \centering
    \vspace{-2.0mm}
    \fontsize{8}{9}\selectfont
    \renewcommand{\arraystretch}{0.85}
    \setlength{\tabcolsep}{2pt}
       \caption{Comparison with training-based method (\ie, Look-Back~\cite{yang2025look}) across benchmarks using InternVL-3.5-8B.}
    \vspace{-2.5mm}
    \begin{tabular}{lccc}
        \toprule
    Method & MathVista~\citep{lumathvista} & MathVison~\citep{wang2024measuring} & MM-Star~\citep{chen2024we}\\ 
       \midrule
        ST & 68.1 & 39.2 & 57.2 \\
        Look-Back~\cite{yang2025look} & 80.8 & 44.2 & 63.7 \\
        \name & 79.3 & 44.6 & 63.1 \\
        Look-Back~\cite{yang2025look}+\name & \textbf{83.1} & \textbf{48.2} & \textbf{66.0} \\
        \bottomrule
    \end{tabular}
    \vspace{-2.5mm}
    \label{tab:lookback_allbench}
\end{table}

\section{Discussion}
\label{sec:discussion}

\noindent\textbf{Understanding the importance of relevance and diversity.} As formalized in Equation~\ref{eq:ours_decomp}, our objective decomposes into two terms: a relevance term that measures the alignment of visual tokens with the current reasoning step $z_k$, and a diversity term that ensures a broad coverage of the visual content. We conduct an ablation study to assess the individual importance of each term. Table~\ref{tab:ablation} presents results on InternVL-3.5-8B across three benchmarks. We observe that the full objective, which jointly optimizes both terms, outperforms variants that use only relevance or only diversity. Notably, using relevance alone results in substantial performance degradation, demonstrating the importance of diversity in selecting effective visual tokens. 

\newcolumntype{?}{!{\vrule width 1pt}}
\newcolumntype{a}{>{\columncolor{myblue}}c}
\newcommand{\cmark}{\ding{51}}%
\newcommand{\xmark}{\ding{55}}%
\begin{table}[!t]
\caption{\small \textbf{Importance of relevance and diversity.} Ablation study to understand the individual importance of the relevance term and diversity term in our scoring function (Equation~\ref{eq:ours_decomp}). We evaluate on InternVL-3.5-8B across all three benchmarks. Accuracy values are reported in percentage (\%). Best results are highlighted in \textbf{bold}.  }
\vspace{-2mm}
\label{tab:ablation}
      \centering
        \resizebox{\columnwidth}{!}{%
        \begin{tabular}{cc|ccccc}
        \toprule
    Relevance &  Diversity & MathVista~\citep{lumathvista} & MathVison~\citep{wang2024measuring} & MM-Star~\citep{chen2024we}\\ 
       \midrule
     
  \cmark & \xmark &  75.6  &  43.3 & 61.0\\
  \xmark & \cmark &  77.4  &  42.9 & 62.8\\
  \midrule
\rowcolor{myblue}  \cmark & \cmark  & \textbf{79.3} & \textbf{44.6} & \textbf{63.1} \\
      \bottomrule
    \end{tabular}%
        }
 \vspace{-2mm}
\end{table}

\noindent\textbf{Analysis of entropy threshold.} We analyze the impact of the entropy threshold $\delta_{\mathrm{entropy}}$, which determines when to terminate reasoning. 
Figure~\ref{fig:ablation}(a) presents the relationship between accuracy and $\delta_{\mathrm{entropy}}$ on {MathVista} using {InternVL3.5-8B}.
While the overall accuracy remains relatively stable across different thresholds, we observe that $\delta_{\mathrm{entropy}} = 0.25$ consistently yields the highest accuracy across all token budgets.
A smaller threshold (e.g., 0.15) requires the model to achieve higher confidence (lower entropy) before stopping, leading to excessive reasoning steps that may introduce overthinking without accuracy gains. Conversely, a larger threshold (e.g., 0.30) can lead to premature termination when the model has insufficient confidence, resulting in under-reasoning and incorrect answer. The same trend is observed for the other two models, Qwen3-VL-8B and  SAIL-VL2, demonstrating that this configuration generalizes well across model scales.

\noindent\textbf{Analysis of token budget.}
We next analyze the effect of the token budget $m$, which controls the number of visual tokens selected at each reasoning step.
Figure~\ref{fig:ablation}(b) shows accuracy as a function of $m$.
Accuracy increases from 76.1\% to 79.2\% as $m$ grows from 20\% to 30\%, but further increasing $m$ to 40\% yields no additional gains.
A similar trend is observed across other models, indicating that \name remains robust to different token-budget configurations.
Based on these findings, we set $m = 30\%$ and $\delta_{\mathrm{entropy}} = 0.25$ for all main experiments, as this configuration achieves a well-balanced trade-off between accuracy and efficiency.

\noindent\textbf{Qualitative evaluation of \name.} To provide deeper insight into how \name maintains visual grounding during reasoning, we visualize attention patterns before and after visual refocusing in Figure~\ref{fig:visualization}. We present examples from MathVista using InternVL-3.5-8B, where we visualize attention maps across reasoning steps. In the initial reasoning phase without visual refocusing, the attention maps exhibit diffuse and noisy patterns, with the model attending broadly across irrelevant image regions while failing to focus on task-critical visual elements. After applying \name's selective visual token reinjection, the attention patterns become substantially more focused and coherent, concentrating on the objects and regions directly relevant to solving the problem.

\section{Conclusion}
\label{sec:conclusion}

\vspace{-1.6mm}
We study how to preserve visual grounding during extended test-time reasoning in multimodal large reasoning models, which tend to overrely on textual priors as the reasoning trace length increases.
We propose \name, a training-free framework that dynamically reinjects visual information throughout reasoning.
Our method selects a compact subset of visual tokens using a DPP-based objective and adaptively terminates reasoning via an entropy-based criterion.
Across three visual reasoning benchmarks, under any fixed test-time compute budget, \name delivers superior accuracy while remaining efficient by selectively reinjecting visual tokens.

{
    \small
    \bibliographystyle{ieeenat_fullname}
    \bibliography{main}
}
\clearpage\newpage

\appendix
\section*{Appendix}

\section{Limitations}

Although \name provides consistent improvements across a variety of visual-reasoning benchmarks and model architectures, it introduces additional computational overhead. In particular, applying DPP-based token selection at every reasoning step increases inference latency compared to standard decoding. This accuracy–latency trade-off is inherent to test-time scaling methods; however, our approach offers higher accuracy for a given computational budget than existing alternatives.


\section{Software and Hardware}

We run all experiments with Python 3.10.18,
PyTorch 2.7.0, and Transformers 4.55.0. For all experimentation, we use four Nvidia A10G GPUs.

\section{Baselines and Implementation Details.} For evaluation, we compare \name with two baselines: (1) standard thinking (Equation~\ref{eq:standard_thinking}), where the model generates a single reasoning trace without additional intervention, and (2) textual self-reflection~\citep{muennighoff2025s1}, which extends reasoning through text-only reflection without visual refocusing. Based on ablations (Section~\ref{sec:experiments}), we set the adaptive stopping entropy threshold $\delta_{\text{entropy}}=0.25$, the visual token budget $m = \lfloor 0.3 |\mathcal{V}| \rfloor$ (i.e., 30\% of the total visual tokens $\mathcal{V}$), and the maximum reasoning steps $K_{\text{max}}=10$.

\section{Evaluation Criteria.} To evaluate reasoning performance, we report the accuracy of each model on the test set of each dataset. Specifically, for each image-text input $x_{\text{input}} = [I, T] \in \mathcal{D}^{\text{test}}$, the model first generates a thinking trace $\tau$, followed by the final answer $y$. The accuracy metric is defined as: $\mathbb{E}_{x_{\text{input}}\sim\mathcal{D}^{\text{test}}, \tau\sim\pi_\theta(\cdot\mid x_{\text{input}}),y\sim\pi_\theta(\cdot \mid x_{\text{input}},\tau)}\left[\mathbb{I}\{y=y^*\}\right]$, where $y^*$ is the correct answer.

\noindent\textbf{Datasets \& Models.} To  validate the effectiveness of \name, we conduct experiments on three visual reasoning benchmarks. 
\textbf{MathVista}~\cite{lu2023mathvista} unifies $31$ visual-math datasets covering puzzles, functional plots, and scientific figures to assess diverse mathematical reasoning skills in visual contexts; we use the \textit{testmini} split containing $1,000$ problems.
\textbf{MathVision}~\cite{wang2024measuring} includes $304$ visually grounded math competition problems across $16$ disciplines and five difficulty levels, enabling fine-grained evaluation of visual mathematical reasoning.
\textbf{MM-Star}~\cite{chen2024we} is a human-curated benchmark of $1,500$ vision-dependent questions designed to assess six core multimodal capabilities across $18$ detailed axes, including perception, spatial reasoning, and commonsense understanding. We evaluate \name on three state-of-the-art MLRMs—InternVL3.5-8B~\cite{wang2025internvl3}, SAIL-VL2-Thinking~\cite{yin2025sail}, and Qwen-3-VL-8B-Thinking~\cite{Qwen2.5-VL}—all evaluated in their reasoning (“thinking”) mode, which generates explicit reasoning traces before producing the final answer.

\section{Additional Results}

\begin{figure*}
    \centering
    \includegraphics[width=0.9\textwidth]{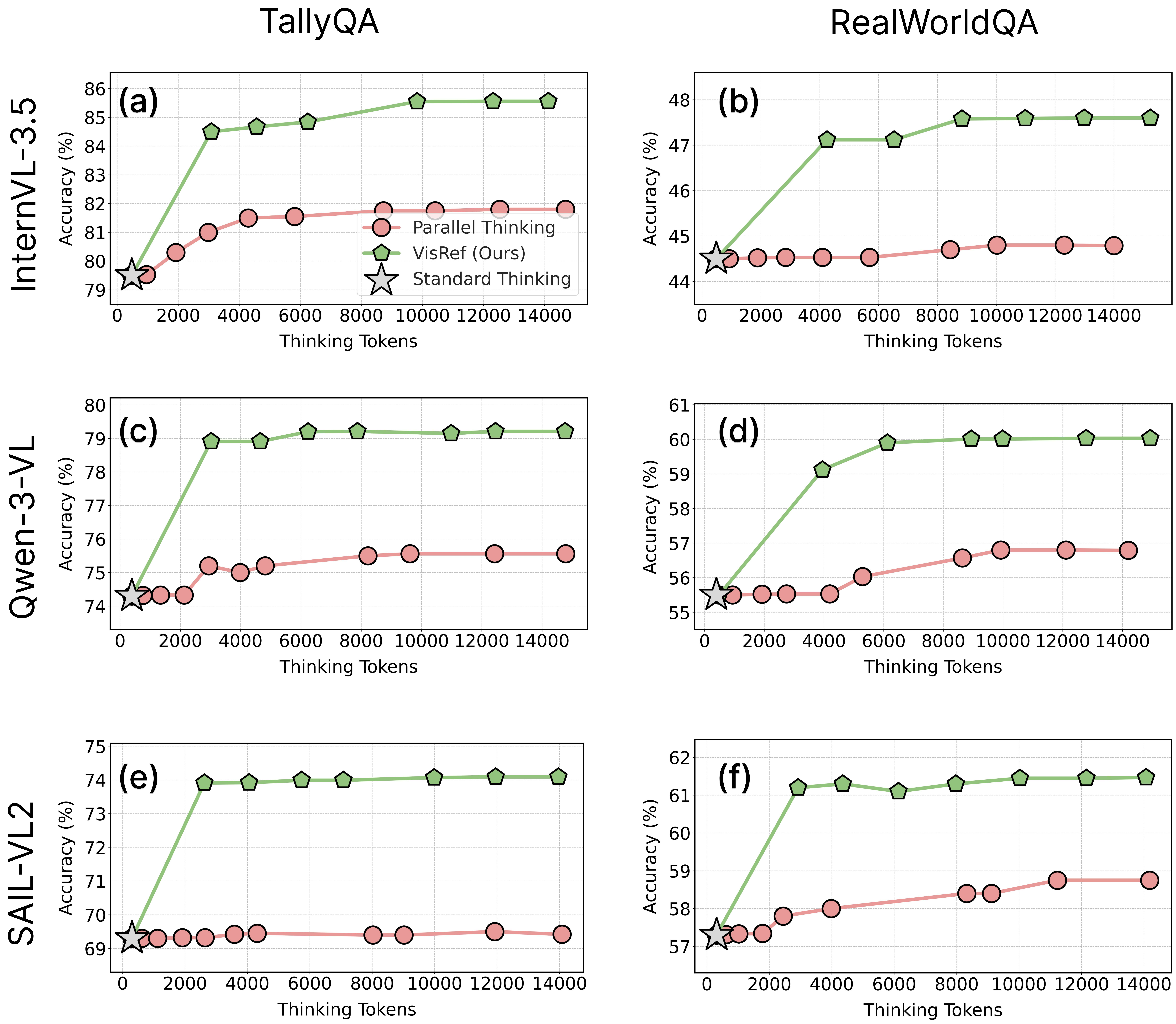}
    \vspace{-2mm}
\caption{ \textbf{Test-time scaling of \name.} We evaluate the test-time scaling behavior of \name by generating multiple parallel visual-integrated reasoning chains under a fixed token budget. Results are shown across two benchmarks (TallyQA, and RealWorldQA) and three MLRMs: InternVL-3.5-8B (first row), Qwen-3-VL-8B (second row), and SAIL-VL2 (third row). The star marker (\ding{73}) denotes standard thinking—the baseline with no additional test-time compute. Parallel thinking~\citep{ghosal2025does, wang2022self} generates multiple parallel chains-of-thought without visual refocusing. Across all models and benchmarks, \name consistently achieves superior accuracy for any given computational budget.  }
\label{fig:parallel_results_additional}
\vspace{-1mm}
\end{figure*}

In the main paper (Section 5), we evaluated \name on three visual reasoning benchmarks. Here, we extend our evaluation to two additional benchmarks: TallyQA~\citep{acharya2019tallyqa} and RealWorldQA~\citep{xai2024realworldqa}.
TallyQA~\citep{acharya2019tallyqa} focuses on complex counting tasks in visual scenes, requiring models to identify and enumerate multiple objects while maintaining spatial awareness. This benchmark is particularly challenging as it tests whether models can preserve precise visual grounding throughout iterative counting processes. RealWorldQA~\citep{xai2024realworldqa} is designed to evaluate real-world visual understanding using over 700 images, including anonymized vehicular footage and diverse real-world scenes. This benchmark tests models' abilities to reason about authentic, uncurated visual scenarios encountered in everyday settings.
Table~\ref{tab:additional_results_adaptive} presents results on these benchmarks. On TallyQA, \name achieves substantial improvements across all three models, with gains of 5.1\%, 4.6\%, and 5.4\% for InternVL3.5-8B, Qwen3-VL-8B, and SAIL-VL2-8B respectively, compared to standard thinking. Similarly, RealWorldQA shows consistent improvements ranging from 2.6\% to 3.9\% across the model suite, demonstrating \name's effectiveness on real-world visual understanding tasks.
Figure~\ref{fig:parallel_results_additional} illustrates the test-time scaling behavior of \name across both additional benchmarks using three MLRMs: InternVL-3.5-8B (top row), Qwen3-VL-8B (middle row), and SAIL-VL2-8B (bottom row). The star marker (\ding{73}) indicates the baseline with no additional test-time compute (standard thinking), while successive circles represent increasing test-time token budgets. We compare \name against parallel thinking~\citep{ghosal2025does, wang2022self}, which samples multiple text-only reasoning trajectories without visual refocusing. Across all benchmarks and models, \name consistently achieves superior accuracy for any given computational budget.

\begin{table}[!t]
\centering
\caption{
\textbf{Evaluation on additional visual reasoning benchmarks.}
We evaluate \name across three visual reasoning benchmarks: {TallyQA}, and {RealQA}. To ensure a fair comparison, all methods adopt the adaptive stopping criterion described in Section 4.1.2.  
For brevity, we denote \textit{Standard Thinking} as \textbf{ST}, and \textit{Textual Self-Reflection}~\citep{muennighoff2025s1} as \textbf{TSR}.  
All results are reported in accuracy (\%), and the numbers in parentheses indicate the performance gain over the ST baseline.
}
\label{tab:additional_results_adaptive}
\resizebox{\linewidth}{!}{
\begin{tabular}{l|l|cc}
\toprule
\multirow{1}{*}{Model} & \multirow{1}{*}{Method} 
& {\textbf{TallyQA}} 
& {\textbf{RealWorldQA}}  \\
\midrule

\multirow{3}{*}{{InternVL3.5-8B}} 
& ST (Baseline) & 79.4  & 44.6  \\
& TSR~\citep{muennighoff2025s1} & 79.6  & 44.9   \\
& \cellcolor{myblue!55}\textbf{VisRef (Ours)} 
  & \cellcolor{myblue!55}\textbf{84.5 (\textcolor{ForestGreen}{+5.1})} 
  & \cellcolor{myblue!55}\textbf{47.2 (\textcolor{ForestGreen}{+2.6})}  \\
\midrule

\multirow{3}{*}{{Qwen3-VL-8B}} 
& ST (Baseline) & 74.3  & 55.4  \\
& TSR~\citep{muennighoff2025s1} & 75.1  & 56.9   \\
& \cellcolor{myblue!55}\textbf{VisRef (Ours)} 
  & \cellcolor{myblue!55}\textbf{78.9 (\textcolor{ForestGreen}{+4.6})}  
  & \cellcolor{myblue!55}\textbf{59.1 (\textcolor{ForestGreen}{+3.7})}  \\
 \midrule

\multirow{3}{*}{{SAIL-VL2-8B}} 
& ST  (Baseline)& 69.3  & 57.3  \\
& TSR~\citep{muennighoff2025s1} & 71.7 & 58.0 \\
& \cellcolor{myblue!55}\textbf{VisRef (Ours)} 
  & \cellcolor{myblue!55}\textbf{73.9 (+\textcolor{ForestGreen}{5.4})} 
  & \cellcolor{myblue!55}\textbf{61.2 (+\textcolor{ForestGreen}{3.9})} \\
\bottomrule
\end{tabular}}
\end{table}

\section{Additional Qualitative Evaluations}

\begin{figure*}[t!]
    \centering
    \includegraphics[width=\linewidth]{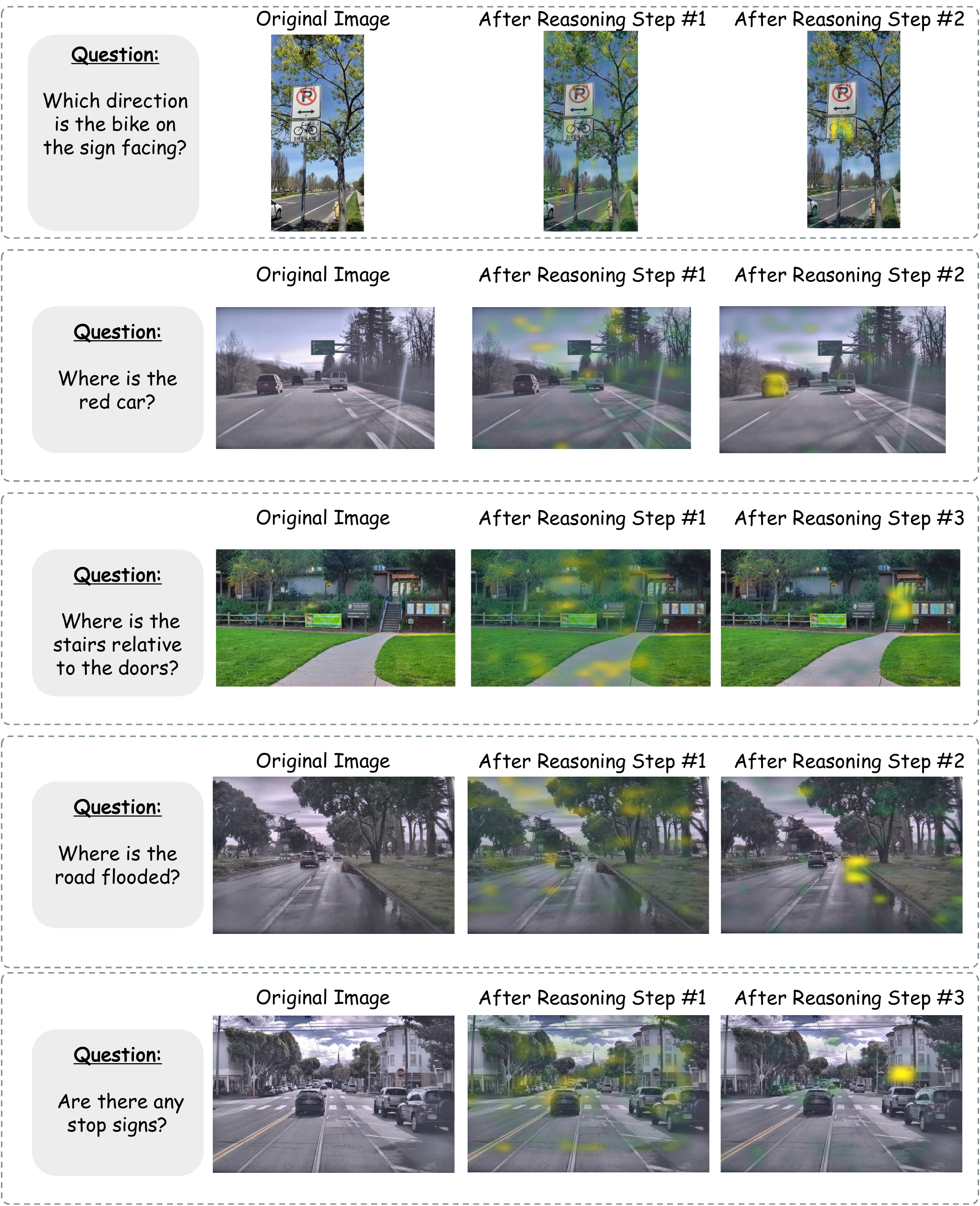}
    \vspace{-5mm}
    \caption{\textbf{Attention Visualization.} 
Attention maps show how \name progressively refocuses on relevant visual regions during multi-step reasoning. Initially, the attention maps are noisy. With visual reinjection, \name  reinforces grounding on task-critical objects, leading to more accurate visual reasoning.
} 
    \vspace{-2.5mm}
    \label{fig:add_visualization}
\end{figure*}

To provide deeper insights into how \name maintains visual grounding during reasoning, we visualize attention patterns before and after visual refocusing in Figure~\ref{fig:add_visualization}. The visualizations use images from the RealWorldQA~\citep{xai2024realworldqa} dataset with the InternVL-3.5-8B model. We observe that after applying \name's visual token reinjection, the attention patterns become substantially more focused on task-relevant regions, confirming that our method effectively counteracts visual token dilution during extended reasoning chains.

\section{Derivation of the Relevance-Diversity Decomposition}

In this section, we provide the complete derivation of Equation 10 from the main paper, which decomposes the log-determinant of the kernel matrix into relevance and diversity terms. Specifically, given the kernel matrix ${L}_k^{{V}_k} \in \mathbb{R}^{|{V}_k| \times |{V}_k|}$ restricted to visual token subset ${V}_k$, the log-determinant can be decomposed as:
\begin{equation}
\log \det({L}_k^{{V}_k}) = \sum_{v_i \in {V}_k} \log(r_i^2) + \log \det(\bar{{L}}_k^{{V}_k})
\end{equation}
where $r_i$ denotes the relevance score of token $v_i$, and $\bar{{L}}_k^{\mathcal{V}_k}$ is the normalized diversity kernel.

\vspace{0.2cm}
\noindent\textbf{Derivation.}  For any selected subset ${V}_k = \{v_1, \ldots, v_m\} \subseteq \mathcal{V}$, the kernel matrix entries are given by:
\begin{equation}
[{L}_k^{{V}_k}]_{ij} = L_k(v_i, v_j) = \phi_k(v_i)^\top \phi_k(v_j) = v_i^\top {M}_k v_j
\end{equation}
where $\phi_k(v) = {M}_k^{1/2} v$ projects visual token $v$ into the textual reasoning subspace defined by ${M}_k = \sum_{j=1}^{T_k} z_k^{(j)}(z_k^{(j)})^\top$.

The relevance of token $v_i$ to the current reasoning state $z_k$ is measured by:
\begin{equation}
r_i^2 = \|\phi_k(v_i)\|_2^2 = v_i^\top {M}_k v_i = \sum_{j=1}^{T_k} (v_i^\top z_k^{(j)})^2
\end{equation}
This quantity captures the alignment between visual token $v_i$ and the textual context, with $r_i^2 = [{L}_k^{{V}_k}]_{ii}$. Next, we introduce the normalized kernel $\bar{{L}}_k^{{V}_k}$ with entries:
\begin{equation}
[\bar{{L}}_k^{{V}_k}]_{ij} = \frac{[{L}_k^{{V}_k}]_{ij}}{r_i r_j} = \frac{\phi_k(v_i)^\top \phi_k(v_j)}{\|\phi_k(v_i)\|_2 \|\phi_k(v_j)\|_2}
\end{equation}
Note that $[\bar{{L}}_k^{{V}_k}]_{ii} = 1$ for all $i$, representing normalized correlations between tokens. Let ${D}_{{V}_k} = \text{diag}(r_1, \ldots, r_m)$. The kernel matrix can then be factorized as:
\begin{equation}
{L}_k^{{V}_k} = {D}_{{V}_k} \bar{{L}}_k^{{V}_k} {D}_{{V}_k}
\end{equation}

Thus, we can write each element as: $[{D}_{{V}_k} \bar{{L}}_k^{{V}_k} {D}_{{V}_k}]_{ij} = r_i \cdot \frac{[{L}_k^{{V}_k}]_{ij}}{r_i r_j} \cdot r_j = [{L}_k^{{V}_k}]_{ij}$. Applying the multiplicative property of determinants:
\begin{equation}
\det({L}_k^{{V}_k}) = \det({D}_{{V}_k})^2 \det(\bar{{L}}_k^{{V}_k}) = \left(\prod_{v_i \in {V}_k} r_i^2\right) \det(\bar{{L}}_k^{{V}_k})
\end{equation}

Finally, taking the natural logarithm yields:
\begin{equation}
\log \det({L}_k^{{V}_k}) = \underbrace{\sum_{v_i \in {V}_k} \log(r_i^2)}_{\text{relevance term}} + \underbrace{\log \det(\bar{{L}}_k^{{V}_k})}_{\text{diversity term}}
\end{equation}

The first term aggregates individual token relevances to the reasoning context, while the second term, through the normalized kernel determinant, penalizes redundancy and encourages diverse visual coverage. This decomposition provides theoretical justification for why maximizing $\log \det({L}_k^{{V}_k})$ naturally balances both objectives.

\section{Computational cost analysis.}

Table~\ref{tab:latency_mathvista} reports detailed latency measurements (on 1 H100 GPU) on the Mathvista dataset using InternVL-3.5-8B. On average, our DPP-based token selection adds only $0.5$ secs of overhead compared to Textual self-reflection (TSR), and $1.1$ secs compared to standard thinking (ST). Note that ST does not include self-reflection, so it is faster than others. The efficiency of \name stems from greedy approximation of Eq.~11.

\begin{table}[t]
    \centering
    \fontsize{9}{9}\selectfont
    \renewcommand{\arraystretch}{0.95}
    \setlength{\tabcolsep}{15pt}
    \begin{tabular}{lc}
        \toprule
        Method & Time \\
        \midrule
        ST & 7.1s \\
        TSR & 7.7s \\
        Look-Back~[1] & 7.6s \\
        \name & 8.2s \\
        \bottomrule
    \end{tabular}
    \caption{Latency per prompt on MathVista.}
    \label{tab:latency_mathvista}
\end{table}
\section{Generalization Across Model Scales}
\label{app:scale_generalization}

We study whether \name consistently improves multi-step visual reasoning as the backbone model scales.
Specifically, we evaluate InternVL models spanning 1B, 2B, and 8B parameters under the same decoding setup and token budget.
Table~\ref{tab:scale_mathvista} shows that \name yields gains over both Standard Thinking (ST) and Textual Self-Reflection (TSR) at every scale:
for the 1B model, \name improves accuracy from 46.1\% (ST) and 48.5\% (TSR) to 52.0\%;
for 2B, it increases accuracy from 52.9\%/53.7\% to 58.1\%;
and for 8B, it improves performance from 68.1\%/73.9\% to 79.3\%.
These results suggest that the benefit of visual refocusing is not confined to a particular parameter regime, but instead persists from small to larger models, indicating that \name effectively counteracts visual token dilution during extended reasoning across model capacities.

\begin{table}[t]
    \centering
    \fontsize{9}{9}\selectfont
    \renewcommand{\arraystretch}{0.95}
    \setlength{\tabcolsep}{2.0pt}
    \begin{tabular}{lccc}
        \toprule
        Model & ST & TSR & \name \\
        \midrule
        InternVL-1B & 46.1 & 48.5 & \textbf{52.0} \\
        InternVL-2B & 52.9 & 53.7 & \textbf{58.1} \\
        InternVL-8B & 68.1 & 73.9 & \textbf{79.3} \\
        \bottomrule
    \end{tabular}
    \caption{Accuracy (\%) across model scales on MathVista.}
    \label{tab:scale_mathvista}
\end{table}

\section{Random Sampling Baseline}
\label{app:random_sampling}

We further verify that the improvement of \name is not merely due to selecting \emph{any} subset of visual tokens under a fixed budget.
To this end, Table~\ref{tab:selection_internvl} compares three selection strategies on InternVL-8B: (i) \emph{Random} selection, (ii) \emph{Relevance-only} selection that greedily keeps the most text-aligned tokens, and (iii) our DPP-based selection that jointly optimizes relevance and diversity.
Random selection performs close to the ST baseline and is substantially worse than \name across all benchmarks, indicating that naive token subsampling fails to preserve the visual evidence needed for multi-step reasoning.
Relevance-only selection improves over random sampling, but it remains consistently below DPP (Ours), suggesting that selecting only the most aligned tokens can still be redundant (e.g., repeatedly focusing on similar regions) and may miss complementary evidence elsewhere in the image.
By explicitly encouraging diversity in addition to relevance, DPP (Ours) achieves the best results, supporting our claim that balancing relevance and diversity is essential for effective visual refocusing under tight token budgets.

\begin{table}[t]
    \centering
    \fontsize{9}{9}\selectfont
    \renewcommand{\arraystretch}{0.75}
    \setlength{\tabcolsep}{0.9pt}
    \begin{tabular}{lccc}
        \toprule
        Selection & MVista & MVision & MM-Star \\
        \midrule
        Random & 67.3 & 40.8 & 57.3 \\
        Relevance-only & 75.6 & 43.3 & 61.0 \\
        DPP (Ours) & \textbf{79.3} & \textbf{44.6} & \textbf{63.1} \\
        \bottomrule
    \end{tabular}
    \caption{Token selection strategies (InternVL-8B).}
    \label{tab:selection_internvl}
\end{table}

\section{Weighted Version of Eq.~10}
\label{app:lambda_weighted}

We additionally experimented with a weighted objective of the form
$\lambda \cdot \text{relevance} + (1{-}\lambda) \cdot \text{diversity}$.
As shown in Table~\ref{tab:lambda_ablation}, performance peaks at $\lambda{=}0.5$ across both MathVista and MathVision,
indicating that balancing relevance and diversity is important in practice.
This result supports our default (unweighted) formulation in the main paper, where the two terms contribute equally.

\begin{table}[t]
    \centering
    \fontsize{8}{9}\selectfont
    \renewcommand{\arraystretch}{0.75}
    \setlength{\tabcolsep}{6pt}
    \begin{tabular}{lccccc}
        \toprule
        $\lambda$ & 0.0 (Div) & 0.25 & 0.5 (Ours) & 0.75 & 1.0 (Rel) \\
        \midrule
        MVista  & 71.2 & 76.8 & \textbf{79.3} & 77.4 & 75.6 \\
        MVision & 41.5 & 43.1 & \textbf{44.6} & 44.2 & 43.3 \\
        \bottomrule
    \end{tabular}
    \caption{Effect of $\lambda$ weighting (InternVL-8B).}
    \label{tab:lambda_ablation}
\end{table}

\end{document}